\documentclass[pdflatex,sn-mathphys-num]{sn-jnl}

\usepackage{graphicx}%
\usepackage{multirow}%
\usepackage{multicol}
\usepackage{amsmath,amssymb,amsfonts}%
\usepackage{amsthm}%
\usepackage{mathrsfs}%
\usepackage[title]{appendix}%
\usepackage{xcolor}%
\usepackage{textcomp}%
\usepackage{manyfoot}%
\usepackage{booktabs}%
\usepackage{algorithm}%
\usepackage{algorithmicx}%
\usepackage{algpseudocode}%
\usepackage{listings}%

\usepackage{lmodern} 
\usepackage{microtype}
\usepackage{float}

\usepackage{amssymb,amsmath}
\usepackage{lmodern}

\usepackage{longtable}
\usepackage{arydshln}

\usepackage{caption}
\usepackage{subcaption}

\theoremstyle{thmstyleone}%
\theoremstyle{thmstyletwo}%

\theoremstyle{thmstylethree}%

\begin{document}

\title[]{A Penalized Shared-parameter Algorithm for Estimating Optimal Dynamic Treatment Regimens}


\author*[1,2]{\fnm{Palash} \sur{Ghosh}}\email{palash.ghosh@iitg.ac.in}

\author[2]{\fnm{Xinru} \sur{Wang}}\email{xinruwang@u.duke.nus.edu}

\author[1]{\fnm{Trikay} \sur{Nalamada}}\email{trikayn@gmail.com}

\author[1]{\fnm{Shruti} \sur{Agarwal}}\email{the.original.shruti@gmail.com}

\author[3]{\fnm{Maria} \sur{Jahja}}\email{maria@stat.cmu.edu}

\author[2,4,5]{\fnm{Bibhas} \sur{Chakraborty}}\email{bibhas.chakraborty@duke-nus.edu.sg}

\affil*[1]{\orgdiv{Department of Mathematics}, \orgname{Indian Institute of Technology Guwahati}, \orgaddress{ \state{Assam}, \country{India}}}

\affil[2]{\orgdiv{Centre for Quantitative Medicine}, \orgname{Duke-NUS Medical School, National University of Singapore}, \orgaddress{\state{Singapore}, \country{Singapore}}}

\affil[3]{\orgdiv{Department of Statistics}, \orgname{North Carolina State University}, \orgaddress{\state{North Carolina}, \country{USA}}}

\affil[4]{\orgdiv{Department of Statistics and Data Science}, \orgname{National University of Singapore}, \orgaddress{\state{Singapore}, \country{Singapore}}}

\affil[5]{\orgdiv{Department of Biostatistics and Bioinformatics}, \orgname{Duke University}, \orgaddress{\state{Durham, NC}, \country{USA}}}

 



\abstract{A dynamic treatment regimen (DTR) is a set of decision
rules to personalize treatments for an individual using their medical history. The Q-learning-based Q-shared algorithm has been used to develop DTRs that involve decision rules shared across multiple stages of intervention. We show that the existing Q-shared algorithm can suffer from non-convergence due to the use of linear models in the Q-learning setup, and identify the condition under which Q-shared fails. We develop a penalized Q-shared algorithm that not only converges in settings that violate the condition, but can outperform the original Q-shared algorithm even when the condition is satisfied. We give evidence for the proposed method in a real-world application and several synthetic simulations. }




\maketitle

\section{Introduction}
 A dynamic treatment regimen (DTR) refers to a set of decision rules for developing personalized multistage treatment using the patient's covariate and treatment history \citep{murphy03, robins04, Chak_book, Ghosh2020noninfi}. At each stage of the disease, a DTR can recommend the best treatment for an individual based on their medical history up to that time.  In recent times, DTRs have been used in various domains of medical and social sciences to leverage data-driven personalized medicine; for example, in mental health \citep{nahum2015Health_psycho, eden2017field, song2015sparse}, cancer \citep{Zhang12a, kidwell2014smart}, autism \citep{almirall2016Autism}, weight loss \citep{Almirall2014}, education \citep{Walkington2013Education} and others.

In dynamic treatment regimens, a shared decision rule is a common function of the time-varying covariate(s) across stages. For instance, a patient suffering from depression is medicated when the depression score (e.g., Quick Inventory of Depressive Symptomatology (QIDS)) crosses a threshold independent of the duration of diagnosis or treatment \citep{rush2003QUID}. In HIV patients, Interleukin 7 (IL-7) injections are given when the CD4 count falls below a threshold \citep{villain2019adaptive}. Treatments are given to diabetic patients when the HbA1c (Hemoglobin A1c) crosses a threshold \citep{international2009international}. Note that in all three examples, a treatment or intervention is offered based on a specific threshold that does not change over time or stages of the disease. This means the decision rule to offer treatment is \textsf{\textit{shared}} across different stages. A more sophisticated example of a shared decision rule is to inform a cardiovascular patient whether a hypertension medication is recommended at each decision point considering the risk factors like age, blood pressure, cholesterol, high-density lipoprotein, diabetes status, and smoking \cite{zhao2020constructing}. The dependence structure of the outcome variable, the long-term risk of cardiovascular disease, on the risk factors is expected to remain the same (shared) over different stages of the disease.

The Q-shared algorithm, proposed by Chakraborty et al. \cite{Chak_2016}, is a Q-learning-based procedure for estimating shared-parameter DTRs, where the Q-function is modelled via linear regression. Some of the parameters in the stage-specific linear regression models remain the same across the stages to capture the shared property of the decision rules. Thus, we have to estimate fewer parameters in the Q-shared algorithm compared to the similar model in the standard (unshared) Q-learning. Therefore, Q-shared is computationally easier to implement, and the corresponding results may be more acceptable to clinicians when the decision rule to provide treatments remains the same over different stages of the disease. However, Robins \cite{robins04} in his work on simultaneous g-estimation, and Chakraborty et al. \cite{Chak_2016} pointed out that the use of a linear model may lead to the failure of the convergence of the estimated parameters to their true values. This drawback of the Q-shared algorithm can limit its use. We explore the issue of non-convergence under the Q-shared setup and find examples of when the Q-shared method fails. Finally, we propose a shrinkage-based penalized Q-shared algorithm that avoids the non-convergence issue. Song et al. \cite{RuiSong2015} previously used shrinkage methods in Q-learning. However, that was in the context of unshared Q-learning decision rules. Recently, Zhao et al. \cite{zhao2020Shared_Censored} proposed two methods called the censored shared-Q-learning and the censored shared-O-learning for censored data. However, their approaches are not based on shrinkage methods, even though they mentioned the lasso penalty.

In the literature, there are other methods than Q-learning to estimate unshared DTRs. A-learning is an alternative approach to Q-learning \citep{schulte2014q}. It is a doubly robust approach to mitigate selection biases in observational studies. Other regression-based approaches include targeted maximum likelihood \cite{vanderLaanRubin2006} and robust Q-learning \cite{ertefaie2021robust}. Estimation of optimal DTRs can also be conducted using value-search or policy-search methods \citep{tsiatis2019dynamic}. For example, C-learning, a value-search method, considers the problem as a weighted classification problem \cite{zhang2018interpretable}.
Additionally, methods for developing interpretable DTRs using tree structures have been proposed to provide clinically meaningful DTRs \citep{zhou2023estimating}.


In this work, our main contributions are: a) to demonstrate at least two instances where linear models, as Q-functions in the Q-shared algorithm, fail to converge; and b) to propose a penalized Q-shared algorithm that avoids the non-convergence problem and gives better \textit{treatment allocation matching} compared to the original Q-shared algorithm. Therefore, penalized Q-shared algorithm is a better option compared to Q-shared, irrespective of the non-convergence issue. The proposed algorithm can be easily implemented in a shared-parameter estimation problem.  

The rest of the paper is organized as follows. In Section 2, we introduce notations and review the basic (unshared) Q-learning as well as the existing Q-shared algorithm. In Section 3, we illustrate the non-convergence of the Q-shared algorithm. Next, we propose the penalized Q-shared algorithm in Section 4. Section 5 presents a simulation study showing the merits of penalized Q-shared over the basic Q-shared. We apply the proposed penalized Q-shared method on a depression clinical trial dataset in Section 6. Section 7 concludes the paper with a discussion.

\section{Optimal DTR with Shared Parameters: Q-shared Algorithm}
Let us consider a study with $J$ stages. In this study, longitudinal data for a patient is given as
$(O_{1},A_{1},...,O_{J},A_{J},O_{J + 1})$ where \(O_{j}\) are the covariates measured prior to the treatment at the \(j^{\text{th}}\) stage and \(A_{j}\) is the treatment assigned at the \(j^{\text{th}}\) stage, for $j=1,\cdots, J$. The $O_{J+1}$ denotes the end-of-study observation at the end of the $J^{th}$ stage \citep{Chak_book, Chak_2016}. The binary treatment options available at each stage, coded \(\{ t_{1},t_{2}\}\), are randomly assigned with known probabilities (e.g., even randomization). The primary outcome is defined as a function of the patient's history as $Y = g(H_{J + 1})$, where $g(\cdot)$ is a real-valued function and history $H_{j}  = (O_{1},A_{1},\dots,O_{j})$ for \(j\  = \ 1,2,\dots,J+1 \). The objective functions, also known as Q-functions \citep{murphy05b}, are given by
\begin{eqnarray}
Q_{J}(H_{J},A_{J}) &=& \mathbb{E}\lbrack Y|H_{J},A_{J}\rbrack \nonumber\\
Q_{j}(H_{j},A_{j}) &=& \mathbb{E}\lbrack \max_{a_{j+1}} Q_{j + 1}(H_{j + 1},A_{j + 1})|H_{j},A_{j}\rbrack \text{ for } j = J-1,\dots,1  \nonumber
\end{eqnarray}
The optimal DTR is \(d_{j}(h_{j}) = arg\ max_{a_{j}}\ Q_j(h_{j},a_{j})\).
The Q-functions can be estimated by linear regression as \(Q_{j}(H_{j},A_{j};\beta_{j},\psi_{j}) = {\beta_{j}}^{T}H_{j0} + \ ({\psi_{j}}^{T}H_{j1})A_{j}\)
where \(H_{j0}\) is the ``main effect of history'' and \(H_{j1}\) is the
``interaction effect of history''. Hence, the optimal DTR can be expressed
as \(d_{j}(h_{j}) = arg\ max_{a_{\text{j\ }}}Q_{j}(h_{j},a_{j};\beta_{j},\psi_{j})\). An optimal DTR, defined as the vector of decision rules \((d_{1},...,d_{J})\), when
estimated using standard Q-learning, is calculated by recursively moving backward through stages. This setup can be called Q-unshared. Let us now look at the Q-shared algorithm where the decision rule parameters are shared across different stages \citep{Chak_2016}. Here, we have $\psi_{1} = \dots = \psi_{J} = \psi$. The Q-functions can be estimated as
\begin{eqnarray}
Q_{j}(H_{j},A_{j}) = \mathbb{E}\lbrack Y_{j}(\theta_{j + 1})|H_{j},A_{j}\rbrack \text{ for } j = 1,2,...,J - 1, \nonumber
\end{eqnarray}
where \({\theta_{j}}^{T} = ({\beta_{j}}^{T},{\psi^{T}}_{})\), \(\beta_{j}\)s are unshared parameters while $\psi$ is the shared parameter, and
\begin{eqnarray}
Y_{j}(\theta_{j + 1}) = {\beta^{T}_{j + 1}}H_{j + 1,0} + \max\{t_1 \times {\psi^{T}_{j + 1}}H_{j + 1,1}, t_2 \times {\psi^{T}_{j + 1}}H_{j + 1,1}\} \nonumber
\end{eqnarray}
is the population-level \(j^{\text{th}}\) stage pseudo-outcome. The pseudo-outcome is the potential outcome if the patient would obtain optimal treatments at later time stages \cite{Chak_book}. For $\{t_{1},t_{2}\}= \{-1,1\}$, $\max\{t_1 \times {\psi^{T}_{j + 1}}H_{j + 1,1}, t_2 \times {\psi^{T}_{j + 1}}H_{j + 1,1}\} = |\psi^{T}_{j + 1}H_{j + 1,1}|$. The Q-function for the $J^{th}$ stage remains the same. \(J\) regression equations can be written as \(Q_{j} = \mathbf{Z}_{j}\theta_{j}\) for \(j\  = \ 1,\dots,J\) where $\mathbf{Z}_{j} = ({H_{j0}}^{T},{H_{j1}}^{T}A_{j})$ and \({\theta_{j}}^{T} = ({\beta_{j}}^{T},{\psi^{T}}_{})\).
Combining data from J stages, we get,
$$
\begin{pmatrix}
Y\\
Y_{J-1}(\theta_J)\\
\vdots\\
Y_1(\theta_2)\\\
\end{pmatrix} \ = \ 
\begin{pmatrix}
\mathbf{Z}_J^{(\beta_J)} & \mathbf{0} & \cdots & \mathbf{0} & \mathbf{Z}_J^{\psi}\\
\mathbf{0} & \mathbf{Z}_{J-1}^{\beta_{J-1}} & \cdots & \mathbf{0} & \mathbf{Z}_{J-1}^{\psi}\\
\vdots & \vdots & & \vdots & \vdots\\
\mathbf{0} & \mathbf{0} & \cdots & \mathbf{Z}_{1}^{\beta_{1}} & \mathbf{Z}_{1}^{\psi}
\end{pmatrix}
\times
\begin{pmatrix}
\beta_{J}\\
\beta_{J-1}\\
\vdots\\
\beta_1\\
\psi
\end{pmatrix}
+
\varepsilon
$$
which can be written as \(Y^{*}(\theta) = \mathbf{Z}\theta  + \varepsilon\) with $\theta^{T} = ({\beta_{J}}^{T},...,{\beta_{1}}^{T},\psi^{T})$ and 
\(\varepsilon\) as the error term. To estimate $\theta$, we wish to minimize 
\begin{eqnarray}
J(\theta)\  = \ ||Y^{*}(\theta) - \mathbf{Z}\theta||^{2}. \nonumber
\end{eqnarray}
This poses a problem because \(Y^{*}(\theta)\) involves the
unknown \(\theta\) term that includes shared parameters. In the absence of a closed-form solution, Chakraborty et al. \cite{Chak_2016} proposed an iterative Q-shared algorithm by using the \emph{Bellman residual}, $Y^{*}(\theta) - \mathbf{Z}\theta$, instead of directly minimizing $J(\theta)$. They took the score function as $S_{\theta}(\mathbf{Z}) = \frac{\partial}{\partial\theta\ }((\mathbf{Z}\theta)^{T}) = \mathbf{Z}^{T}$, and formulated the estimating equation as $S_{\theta}(\mathbf{Z})\  \times (Y^{*}(\theta) - \mathbf{Z}\theta)\  = 0$. Algorithm \ref{Q-shared_algo} describes the steps to estimate parameters using the Q-shared algorithm. 



\begin{algorithm}[ht]
	\caption{Q-shared algorithm:}
	\begin{enumerate}
		\def\labelenumi{(\arabic{enumi})}
		\item
		Set starting value of \(\theta\) (one can estimate from data, or guess); define it as ${\hat{\theta}}^{(0)^{T}} = ({{\hat{\beta}}_{J}}^{(0)^{T}},\dots,{{\hat{\beta}}_{1}}^{(0)^{T}},{\hat{\psi}}^{(0)^{T}})$
		
		\item For $k = 0,1,2,\dots$, at the \((k + 1)^{\text{th}}\) iteration:
		\begin{enumerate}
			\def\labelenumii{\alph{enumii}.}
			\item[a)] Formulate the vector \(Y^{*}({\widehat{\theta}}^{(k)})\).
			\item[b)] To obtain ${\hat{\theta}}^{(k+1)}$, solve the estimating equation
			\(\mathbf{Z}^{T}(Y^{*}({\hat{\theta}}^{(k)}) - \mathbf{Z}{\theta}) = 0\)
			for $\theta$.
		\end{enumerate}
		\item
		Repeat (2a) and (2b) until convergence. Specifically,  for a prespecified value of $\varepsilon$ and for some $k$,  continue till $||{\hat{\theta}}^{(k + 1)} - {\hat{\theta}}^{(k)}|| < \varepsilon$.
	\end{enumerate}
	\label{Q-shared_algo}
\end{algorithm}

\section{Non-convergence of the Q-shared algorithm}\label{Non-covg_Q-Pena}
The Q-shared algorithm shares similarities with the ubiquitous least squares regression, and we can better ground our understanding of Q-shared through the intuition of least squares. The ordinary least squares estimate for the model $Y \sim {\mathbf{X}}\beta$ is $\hat{\beta} = arg\min_{\beta}||{\mathbf{X}}\beta\  - \ Y||_{2} = ({\mathbf{X}}^{T}\mathbf{X})^{- 1}{\mathbf{X}}^{T}Y$. Let \(\hat{Y}\) be the predicted value for \(Y\), then $\hat{Y} = {\mathbf{X}}\widehat{\beta} = {\mathbf{X}}({\mathbf{X}}^{T}{\mathbf{X}})^{- 1}{\mathbf{X}}^{T}Y = \mathbf{H}Y$, where \(\mathbf{H}\) is the hat matrix. Assuming $({\mathbf{X}}^{T}{\mathbf{X}})^{- 1}$ exists (and thus implying \(\mathbf{H}\) exists) and based on the discussion by Lizotte \cite{lizotte2011convergent}, we can say that Q-shared may fail when the corresponding hat matrix \(\mathbf{H}\) is \textsf{\textit{not}} an \(\infty\)-norm non-expansion. Note that \(\mathbf{H}\) will be \(\infty\)-norm non-expansion if
\begin{eqnarray}
||\mathbf{H}||_{op(\infty)} = \text{max}_{i}{\ \Sigma_{j}\ |\mathbf{H}}_{\text{ij}}|\  \leq 1,
\label{criteria_non-converge}
\end{eqnarray}
where $\mathbf{H}_{\text{ij}}$ is the $(i,j)^{\text{th}}$ element of the hat matrix \(\mathbf{H}\). In other words, for any row of the hat matrix ($\mathbf{H}$), the sum of the absolute values of all the (row) elements ($\mathbf{H}_{\text{ij}}$) is less than one. Note that the $i^{th}$ element of $\hat Y$ can be expressed as a linear combination of the elements of the $Y$, where the weights are given by the $i^{th}$ row of the hat matrix $H$. Thus, (\ref{criteria_non-converge}) restricts the sum of the absolute values of those weights. In general, the \(\infty\)-norm non-expansion described in (\ref{criteria_non-converge}) is defined for a linear operator \cite{lizotte2011convergent}. Gordon \cite{gordon1999approximate} discussed that if an \(\infty\)-norm non-expansion linear operator is used for the approximation in fitted value iteration, the algorithm's (e.g., linear model as the Q-function in Q-shared) convergence is guaranteed from any initial points. Therefore, the convergence of the Q-shared algorithm is not guaranteed during the estimation of parameters if (\ref{criteria_non-converge}) is violated. Robins \cite{robins04} also discussed the non-convergence due to the use of linear models in the context of simultaneous g-estimation. Below, we show two examples where the above Q-shared algorithm fails to converge, indicating a limitation in its applicability.

Consider a three-stage ($J=3$) sequential multiple-assignment randomized trial (SMART) design where responders enter the follow-up phase and exit the study, and non-responders are re-randomized again in the next stage. We use a simulated data set given in Chakraborty et al. \cite{Chak_2016} to show that the Q-shared algorithm fails to converge after a slight modification in the covariates in the same data. The first few rows of the dataset are given in Table \ref{Ex-1-simu_data}, where $Y_1, Y_2$ and $Y_3$ are stage-specific outcomes corresponding to stages 1, 2, and 3, respectively. The primary outcome is defined as $Y_{Primary} = R_1 Y_1 + (1 - R_1 )R_2((Y_1 + Y_2)/2) + (1 - R_1)(1 - R_2)((Y_1 + Y_2 + Y_3)/3)$, where $R_1$ and $R_2$ are response indicators corresponding to stages 1 and 2, respectively (0: non-responder and 1: responder). The stage-specific treatments $A_j, j=1,2,3$ and covariates $O_j, j =1,2,3$ are all binary, coded $\{-1, 1\}$. The total sample size (number of rows) of the data is 300. To apply the Q-shared algorithm, we consider the initial estimates of all the parameters as zeros, i.e., ${{\hat{\theta}}_{}}^{(0)^{}} = ({{\hat{\beta}}_{J}}^{(0)^{T}},...,{{\hat{\beta}}_{1}}^{(0)^{T}},{\hat{\psi}}^{(0)^{T}}) = (0^{T},{...,0}^{T},0^{T})$. Note that, in the Q-shared algorithm, for a specific iteration ($Y^{*}(\theta)$ is known), $X=Z$. In the above data, the Q-shared algorithm converged in 8 iterations. However, we observed that the corresponding $H$ matrix is \textsf{\textit{not}} an $\infty$-norm non-expansion using (\ref{criteria_non-converge}). We see the norm of $\mathbf{H}$ as $ ||\mathbf{H}||_{op(\infty)} = ||\mathbf{X}({\mathbf{X}}^{T}{\mathbf{X}})^{- 1}{\mathbf{X}}^{T}||_{op(\infty)} = 4.52$. The estimated parameters, along with $m$-out-of-$n$ bootstrap variances and 95\% confidence interval (CI), are presented in Table \ref{results_Q-shared-converged}. Note that we have used the $m$-out-of-$n$ bootstrap to restore bootstrap consistency in the presence of non-regularity \citep{ chak_laber12, Chak_2016}.

\begin{table}[ht]
	\centering
	\caption{First few rows of the data set used by \cite{Chak_2016}.}
	\label{Ex-1-simu_data}
	\begin{tabular}{rrrrrrrrrrrrr}
		\hline
		& $Y_1$ & $Y_2$ & $Y_3$ & $Y_{Primary}$ & $A_1$ & $A_2$ & $A_3$ & $O_1$ & $O_2$ & $O_3$ & $R_1$ & $R_2$\\ 
		\hline
		1 & 1.5448 & 2.857 & 2.230 & 1.544 & -1 & -1 & -1 & -1 & -1 & -1 & 1 & 1\\
		2 & -0.4734 & -0.688 & -0.724 & -0.580 & 1 & 1 & 1  & 1  & 1 & 1 & 0 & 1\\
		3 &  0.4853 & -0.327 & -0.741 & 0.485 & 1 & 1 & -1 & -1 & -1 & 1 &  1 & 1\\
		4 & -0.4224 & 0.547 & -0.377 & -0.084 & 1 & -1 & 1 & -1 & 1 & -1 &  0 &  0\\
		: & : & : & : & : & : & : & : & : & : & : & : & : \\
		\hline
	\end{tabular}
\end{table}

\begin{table}[ht]
	\centering
	\caption{Estimated parameters using the Q-shared algorithm.}
	\label{results_Q-shared-converged}
	\begin{tabular}{cccc}
		\hline
& & \multicolumn{2}{c}{ Using $m$-out-of-$n$ bootstrap} \\ \cmidrule{3-4}
		Shared Parameter & Estimate & Variance & CI \\ 
		\hline
		$\psi_0$ & -0.0974 & 0.0061 & (-0.2156, 0.1040) \\ 
		$\psi_1$ & 0.0802 & 0.0024 & (-0.0192, 0.1710) \\ 
		$\psi_2$ & 0.0876 & 0.0055 & (-0.0522, 0.2366) \\ 
		$\psi_3$ & -0.0409 &    0.0101 & (-0.2365, 0.1616) \\ 
		\hline
	\end{tabular}
\end{table}

\subsection*{Example 1: } 
In the above simulated example (Table \ref{Ex-1-simu_data}), covariates  $O_j, j =1,2,3$ are coded as $\{-1, 1\}$. Now, we modify the data by considering $O_1, O_2, O_3 \in \{ -0.01, 0, 0.01\}$, which follow multinomial distributions with probability $1/3$ for each category. Note that, $ A_1, A_2, A_3 \in \{-1, 1\}$, are unchanged. In this modified data, we applied the Q-shared algorithm and observed that $\psi_1$ fails to converge to one particular value in most of the 1000 $m$-out-of-$n$ bootstrap replications. From different iterations, we notice that most of the iterations converge to different values, which result in the high variance of the estimate of $\psi_1$ as shown in Table \ref{Ex1: results_Q-shared-NON-converged}. We notice that $ ||\mathbf{H}||_{op(\infty)} = ||\mathbf{X}({\mathbf{X}}^{T}{\mathbf{X}})^{- 1}{\mathbf{X}}^{T}||_{op(\infty)} = 4.56$ is greater than 1, and hence convergence is not guaranteed, which we observe in our case. Table \ref{Ex1: results_Q-shared-NON-converged} shows the estimated parameters along with $m$-out-of-$n$ bootstrap variances and 95\% confidence interval (CI). Such a high variance of the estimates of $\psi_1$ (in Table \ref{Ex1: results_Q-shared-NON-converged}) is not useful. It does not provide information about the precise value of the estimate, and hence, we need alternative methods to solve the problem with this particular setting.

\begin{table}[ht]
	\label{table_1}
	\centering
		\caption{Example 1: Estimated parameters using Q-shared algorithm from modified data. }
	\label{Ex1: results_Q-shared-NON-converged}		
		\centering
		\begin{tabular}{cccccc}
			\hline
& & \multicolumn{2}{c}{Using m-out-of-n bootstrap} \\ \cmidrule{3-4}
			Shared Parameter & Estimate & Variance & CI \\
			\hline
			$\psi_0$ & -0.109 & 0.006 & (-0.223, 0.088) \\ 
			$\psi_1$ & 1.153 & 34.737 & (-10.027, 12.707) \\ 
			$\psi_2$ & 0.111 & 0.004 & (-0.019, 0.242) \\ 
			$\psi_3$ & -0.071 & 0.012 & (-0.282, 0.148) \\ 
			\hline
		\end{tabular}
\end{table}

\subsection*{Example 2:} 
We now again modify the data used in Example 1 by considering $O_1, O_2, O_3 \in \{ -0.01, 0, 0.01\}$,  which follows multinomial distributions with probability for each category being calculated based on the outcome $Y_j$ and the treatment $A_j$ at stage $j$. Specifically, $Z_j = 1+ 0.6 \times Y_j\times A_j + \epsilon_j, \quad \epsilon_j \sim N(0,1)$ and $O_j = -0.01$ if $Z_j \leq 0.6$; $O_j = 0$ if $0.6  < Z_j \leq 1.2$; $O_j = 0.01$ if $1.2  < Z_j$. Following the same procedure as in Example 1, we once again notice that the shared parameter $\psi_1$ converges to different values in most of the 1000 $m$-out-of-$n$ bootstrap replications and results in a high variance, which is a sign of non-convergence. We notice that $||\mathbf{H}||_{op(\infty)} = ||\mathbf{X}({\mathbf{X}}^{T}{\mathbf{X}})^{- 1}{\mathbf{X}}^{T}||_{op(\infty)} = 4.69$. Table \ref{Ex2: results_Q-shared-NON-converged} shows the estimated parameters along with $m$-out-of-$n$ bootstrap variances and 95\% confidence interval (CI). We observe high variance for the estimate of $\psi_1$. The above two examples show that the Q-shared algorithm may not converge always. Therefore, we need an algorithm that can address the non-convergence issue.

\begin{table}[ht]
	\label{table_1}
\centering
		\caption{Example 2: Estimated parameters using Q-shared algorithm from modified data. }
	\label{Ex2: results_Q-shared-NON-converged}		
		\centering
		\begin{tabular}{cccccc}
			\hline
& & \multicolumn{2}{c}{Using m-out-of-n bootstrap} \\ \cmidrule{3-4}
			Shared Parameter & Estimate & Variance & CI \\
			\hline
			$\psi_0$ & -0.065 & 0.002 & (-0.146, 0.014) \\ 
			$\psi_1$ & 40.467 & 12.073 & (34.024, 47.294) \\ 
			$\psi_2$ & 0.019 & 0.003 & (-0.079, 0.129) \\ 
			$\psi_3$ & -0.016 & 0.006 & (-0.175, 0.13) \\ 
			\hline
		\end{tabular}
\end{table}


\section{Penalized Q-shared Algorithm} \label{pena_Q-shared}
We have demonstrated that the Q-shared algorithm fails to converge in at least two examples. Moreover, from the $m$-out-of-$n$ bootstrap replicates, we observe high variances for estimates of the shared parameters.
It prompts us to think about shrinkage methods inside the Q-shared algorithm instead of the linear regression model to control the high variability \citep{hastie2009elements}. In Algorithm \ref{Pena_Q-shared_algo}, we propose the penalized Q-shared algorithm that uses ridge regression as the shrinkage method. 



\begin{algorithm}[ht]
	\caption{Penalized Q-shared algorithm:}
	\begin{enumerate}
		\def\labelenumi{(\arabic{enumi})}
		\item
		Set starting value of \(\theta\) (one can estimate from data, or guess); define it as ${\hat{\theta}}^{(0)^{T}} = ({{\hat{\beta}}_{J}}^{(0)^{T}},\dots,{{\hat{\beta}}_{1}}^{(0)^{T}},{\hat{\psi}}^{(0)^{T}})$
		
		\item For $k = 0,1,2,\dots$, at the \((k + 1)^{\text{th}}\) iteration:
		\begin{enumerate}
			\def\labelenumii{\alph{enumii}.}
			\item[a)] Formulate the vector \(Y^{*}({\widehat{\theta}}^{(k)})\).
			\item[b)] Get, ${\hat{\theta}}^{(k+1)}\  = \ arg\ min{}_{\theta} (||{\mathbf{Z}} {{\theta}} - Y^{*}({\hat{\theta}}^{(k)}) ||_{2} + \lambda || {{\theta}} ||_{2})$ 
		\end{enumerate}
		\item
		Repeat (2a) and (2b) until convergence. Specifically,  for a prespecified value of $\varepsilon$ and for some $k$,  continue till $||{\hat{\theta}}^{(k + 1)} - {\hat{\theta}}^{(k)}|| < \varepsilon$.
	\end{enumerate}
	\label{Pena_Q-shared_algo}
\end{algorithm}

Here, note that we do not have to use an approximate estimating equation involving the \textit{Bellman residual} instead of minimizing the \textit{squared Bellman error} to obtain the $\hat\theta^{(k+1)}$. Chakraborty et al. \cite{Chak_2016} used the approximate estimating equation involving the \textit{Bellman residual} to avoid instability or non-convergence during the estimation process. Here, the ridge regression takes care of those issues through the penalty parameter $\lambda$. Note that $\lambda \geq 0$ controls the extent of shrinkage. A larger value of $\lambda$ indicates a higher amount of shrinkage. The optimal value of $\lambda$ in Algorithm \ref{Pena_Q-shared_algo} (step 2(b)) can be obtained by minimizing the 10-fold cross-validation error \citep{friedman2010regularization} as
\begin{eqnarray}
\hat{\lambda} =\ arg\ min{}_{\lambda}\biggl[ \frac{1}{10} \sum_{v=1}^{10} \frac{1}{n_{S_v}} \sum_{u \in S_v}  (Y_{obs, u}(\lambda) - \hat Y_{Pred,u}(\lambda))^2 \biggr], \nonumber
\end{eqnarray}
where $(S_1, S_2, \cdots, S_{10})$ denote the 10 partitions of the data with the corresponding number of subjects as $(n_{S_1}, n_{S_2}, \cdots, n_{S_{10}})$, respectively. $Y_{obs, u}(\lambda)$ =$ $ $Y_{u}^{*} ({\hat{\theta}}^{(K)})$ and $\hat Y_{Pred,u}(\lambda) = \mathbf{Z}_{u} {\hat{\theta}}^{(K)}$ denote the observed and the predicted values of the primary outcome for the $u^{th}$ subject in a partition based on the final iterative stage $K$ of the penalized Q-shared algorithm, respectively. Note that $\hat{\theta}^{(K)}$ is a function of $\lambda$ and thus known for a given value of $\lambda$. 


It is natural to also consider $L_{1}$ regularization. However, $L_{1}$ regularized regression, or the lasso \citep{hastie2009elements}, is generally applied for the purposes of variable selection or denoising; in the context of Q-shared, we are simply attempting to improve variability in parameter estimates at the cost of bias. This is a task best suited for $L_{2}$ regularization.

\section{Simulation Studies}
The objective of the following simulation study is to show that the proposed penalized Q-shared algorithm performs better than the existing Q-shared algorithm in two aspects. First, the penalized Q-shared algorithm does converge when Q-shared does not. Secondly, when there are no issues of non-convergence, the performance of the penalized Q-shared algorithm is better than the Q-shared with respect to allocating appropriate treatments to patients. In other words, there is no reason not to employ penalized Q-shared over the original Q-shared.

\begin{figure}[ht]
	\centering
	\includegraphics[width=\textwidth]{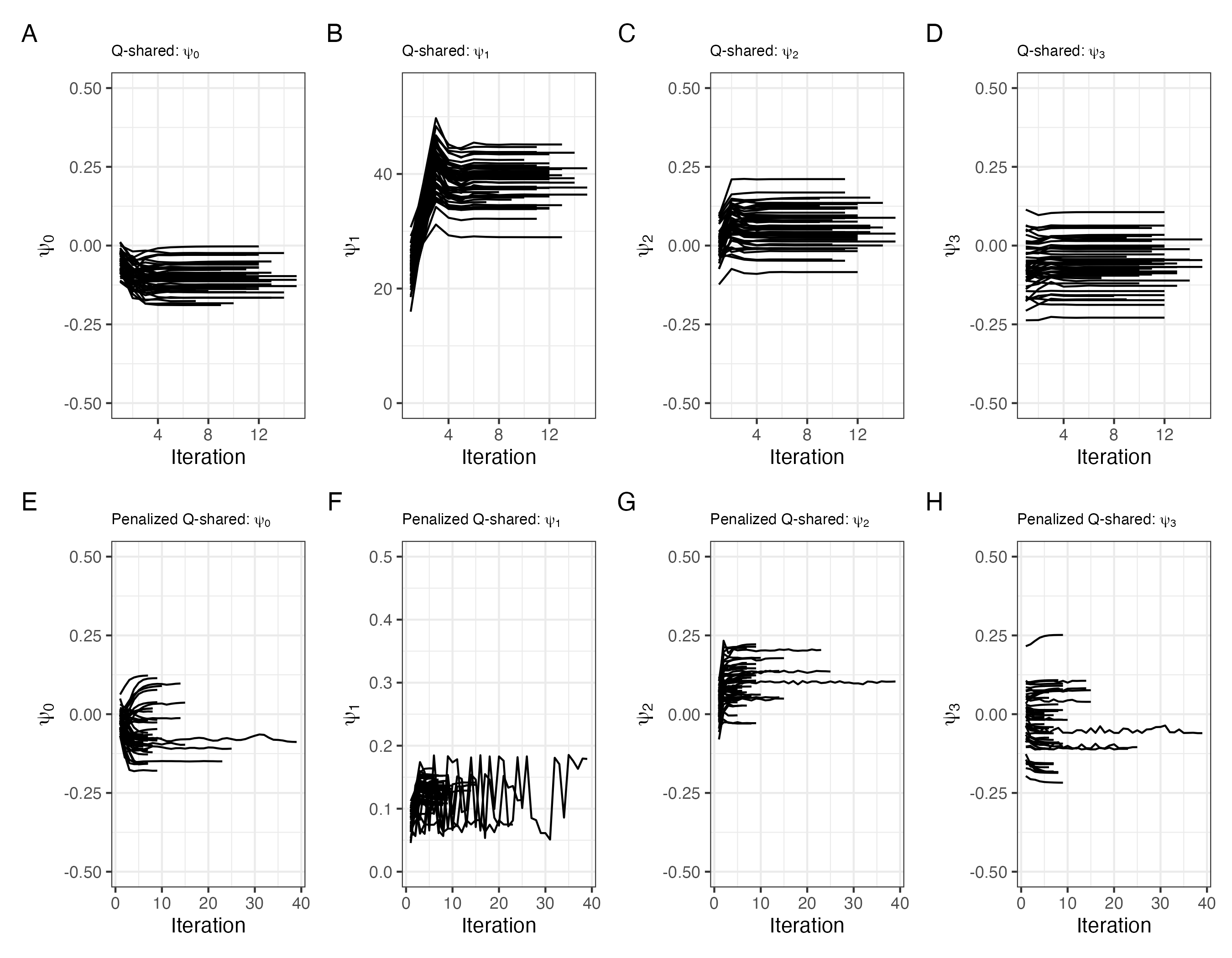}
	\caption{Convergence patterns of Q-shared (top row) and penalized Q-shared (bottom row) based on 50 $m$-out-of-$n$ bootstrap samples. The initial values for all the cases have been set to zero. }
	\label{fig:converge_graphs}
\end{figure}

In Section \ref{Non-covg_Q-Pena}, we have provided two examples where the existing Q-shared algorithm fails to converge. Here, we use the penalized Q-shared algorithm for the same data used for the two examples. Tables \ref{Tab_Ex1:Pena_Qshared} and \ref{Tab_Ex2:Pena_Qshared} show the estimated parameters along with corresponding $m$-out-of-$n$ bootstrap variances and 95\% confidence intervals (CIs) using penalized Q-shared for the Examples 1 and 2, respectively. In both examples, we have used 1000 $m$-out-of-$n$ bootstrap iterations. We can see that in both examples, the estimated parameters have small variances. In other words, when the $O_j, j=1,2,3$ takes values in $\{ -0.01, 0, 0.01\}$ instead of $\{-1, 1\}$, the Q-shared method may not give satisfactory results. It could be a significant issue in a SMART when the covariates are multi-categorical or continuous in nature. In Figure \ref{fig:converge_graphs}, we show the convergence patterns of the Q-shared and the penalized Q-shared algorithms based on 50 $m$-out-of-$n$ bootstrap samples related to Example 2 for all the four parameters $\psi_0, \psi_1, \psi_2$ and $\psi_3$. It is evident from the top row of Figure \ref{fig:converge_graphs} that the Q-shared algorithm fails to converge and exhibits high variability for $\psi_1$, which is consistent from the results reported in Table \ref{Ex2: results_Q-shared-NON-converged}. In contrast, the bottom row of Figure \ref{fig:converge_graphs} indicates a convergence of the penalized Q-shared algorithm with extremely low variances for all four parameters.

\begin{table}[ht]
\centering
		\caption{Estimates of the shared parameters along with their $m$-out-of-$n$ bootstrap variances and confidence intervals generated by the penalized Q-shared algorithm using the same data as in example 1.}
			\label{Tab_Ex1:Pena_Qshared}
		\centering
		\begin{tabular}{cccccc}
			\hline
			& & \multicolumn{2}{c}{Using m-out-of-n bootstrap} \\
			\cmidrule{3-4}
			Shared Parameter & Estimate & Variance & CI \\
			\hline
			$\psi_0$ & -0.099 & 0.005 & (-0.189, 0.080) \\ 
			$\psi_1$ & 0.003 & 0.0001 & (-0.025, 0.027) \\ 
			$\psi_2$ & 0.104 & 0.004 & (-0.023, 0.233) \\ 
			$\psi_3$ & -0.041 & 0.009 & (-0.222, 0.140) \\ 
			\hline
		\end{tabular}
\end{table}


\begin{table}[ht]
\centering
	\caption{Estimates of the shared parameters along with their $m$-out-of-$n$ bootstrap variances and confidence intervals generated by the penalized Q-shared algorithm using the same data as in example 2.}
	\label{Tab_Ex2:Pena_Qshared}
		\centering
		\begin{tabular}{cccccc}
			\hline
			& & \multicolumn{2}{c}{Using m-out-of-n bootstrap} \\
			\cmidrule{3-4}
			Shared Parameter & Estimate & Variance & CI \\
			\hline
			$\psi_0$ & -0.099 & 0.005 & (-0.194, 0.083) \\ 
			$\psi_1$ & 0.131 & 0.0004 & (0.076, 0.166) \\ 
			$\psi_2$ & 0.104 & 0.003 & (-0.008, 0.220) \\ 
			$\psi_3$ & -0.042 & 0.009 & (-0.233, 0.142) \\ 
			\hline
		\end{tabular}

\end{table}


In a shared parameter DTR problem, Chakraborty et al. \cite{Chak_2016} showed that the Q-shared algorithm works better than Q-unshared using simulation. In Table \ref{Tab: allocation-match}, our objective is to illustrate that the penalized Q-shared algorithm performs even better than the original Q-shared algorithm.  We use the same set of examples and the same metrics for comparison, as in Chakraborty et al. \cite{Chak_2016}, to show the differences between the two methods. As mentioned in Section \ref{pena_Q-shared}, the penalized Q-shared algorithm requires initial values of the parameters to start the algorithm. In Table \ref{Tab: allocation-match}, we have considered five different choices of initial values of parameters. First, we have used Q-unshared to estimate $\theta_{j}^{T} = (\beta_{j}^{T},{\psi_{j}^{T}})$ as $\hat\theta_{j}^{(0)T} = (\hat\beta_{j}^{(0)T},{\hat\psi_{j}^{(0)T}}), j=1,\cdots,J$. Five different choices of the initial values of each of the shared parameters are Simple Average (SA): $(1/J)\sum_{j=1}^{J}\hat\psi_{j}^{(0)}$; Inverse Variance Weighted Average (IVWA): $(\sum_{j=1}^{J}\frac{\hat\psi_{j}^{(0)}}{\hat\sigma_{j}^2})/(\sum_{j=1}^{J}\frac{1}{\hat\sigma_{j}^2})$, where the estimated large sample variance of $\hat\psi_{j}^{(0)}$ is $\hat\sigma_{j}^2$; MAX: $\max\{\hat\psi_{1}^{(0)}, \cdots, \hat\psi_{J}^{(0)}\}$; MIN: $\min\{\hat\psi_{1}^{(0)}, \cdots, \hat\psi_{J}^{(0)}\}$; ZERO: all the initial values of the parameters are taken as zero. Note that the initial values of the intercept parameter $\psi_0$ are calculated in similar ways for all the five different choices.

We consider two metrics: the weighted average of the stage-specific allocation matching ($M$) and overall allocation matching ($\Tilde{M}$) over all the stages of the study. By matching, we mean the matching of treatment allocation between the method under consideration and an ``oracle'' method. The oracle method assumes that the Q-functions are correctly specified and the true values of the model parameters are known. Thus, the oracle DTR $d^{\psi_{oracle}}$ can allocate the best treatments over all the stages. To balance the different number of subjects across the different stages, the weighted average of the stage-specific allocation matching is defined as
$M = \frac{\sum_{j=1}^{J} n_jM_j}{\sum_{j=1}^{J} n_j}$, where $n_j$ denotes the number of patients at the $j^{th}$ stage, $M_j = P[d_j^{\hat\psi}(H_j) = d_j^{\psi_{oracle}}(H_j)]$ is the probability of allocation matching with the oracle at stage $j$ and $d_j^{\hat\psi}$ is the estimated DTR from the method under consideration. Similarly, the overall allocation matching is defined as $\Tilde{M} = P[d^{\hat\psi} = d^{\psi_{oracle}}]$, which denotes the probability of allocation matching over all the stages of the study for a subject. 

In this simulation study, we consider a three-stage ($J=3$) SMART design with 300 patients. Recall that, at the end of each stage, nonresponders proceed to the next stage ($j$ to $j+1$), and responders proceed to a follow-up phase and exit the main study. As defined in Section \ref{Non-covg_Q-Pena}, $R_1$ and $R_2$ are the response indicators. Here, we generate $R_1$ and $R_2$ using a Bernoulli distribution with success probability 0.38 and 0.18, respectively. These values are obtained from the sequenced treatment alternatives
to relieve depression (STAR*D) trial; our simulation study follows the same structure as this trial \citep{rush04}. 

The stage-specific outcomes are generated as \citep{Chak_2016},
\begin{eqnarray}
Y_1 = \gamma_1 + \gamma_2 O_1 + \gamma_3 A_1 + \gamma_4 O_1A_1 + \epsilon_1, \nonumber \\
Y_2 = Y_1 + \frac{3}{2}[ \gamma_5 O_2 + \gamma_6 A_2 + \gamma_7 O_2A_2 + \gamma_8 A_1A_2] + \epsilon_2, \nonumber \\
Y_3 = Y_2 + 3[ \gamma_9 O_3 + \gamma_{10} A_3 + \gamma_{11} O_3A_3 + \gamma_{12} A_2A_3 + \gamma_{13} A_1A_2A_3] + \epsilon_3, \nonumber
\end{eqnarray}
where for $j=1,2,3$, the error terms $\epsilon_i \sim N(0,1)$, $A_j$ takes values 1 or -1 with equal probability (0.5); the covariate $O_1$ also takes values 1 or -1 with equal probability, $P(O_2 = 1| O_1, A_1) = 1- P(O_2 = -1| O_1, A_1) = \exp(\delta_{21}O_1 + \delta_{22}A_1)/(1+\exp(\delta_{21}O_1 + \delta_{22}A_1))$ and $P(O_3 = 1| O_1, A_1, O_2, A_2) = 1- P(O_3 = -1| O_1, A_1, O_2, A_2) = \exp(\delta_{31}O_2 + \delta_{32}A_2 + \delta_{33}A_1A_2)/(1+\exp(\delta_{31}O_2 + \delta_{32}A_2 + \delta_{33}A_1A_2))$. As defined in Section \ref{Non-covg_Q-Pena}, the primary outcome is $Y_{Primary} = R_1 Y_1 + (1 - R_1 )R_2((Y_1 + Y_2)/2) + (1 - R_1)(1 - R_2)((Y_1 + Y_2 + Y_3)/3)$. Therefore, above expressions for $Y_1, Y_2$ and $Y_3$ ensure that the final outcome for those who remain till the end ($R_1=R_2=0$) of the third stage is $Y = \gamma_1 + \gamma_2 O_1 + \gamma_3 A_1 + \gamma_4 O_1A_1  + \gamma_5 O_2 + \gamma_6 A_2 + \gamma_7 O_2A_2 + \gamma_8 A_1A_2 +  \gamma_9 O_3 + \gamma_{10} A_3 + \gamma_{11} O_3A_3 + \gamma_{12} A_2A_3 + \gamma_{13} A_1A_2A_3 + \epsilon$, where $\epsilon = (\epsilon_1 + \epsilon_2 + \epsilon_3)/3$. Given the $\gamma$-parameters, the above generative models generate a 3-stage SMART data for the simulation study. From this data, we develop DTRs using the following Q-functions:
\begin{eqnarray}
Q_3 &=& \beta_{30} + \beta_{31}O_1 + \beta_{32}A_1 + \beta_{33}O_1A_1 + \beta_{34}O_2 + \beta_{35}A_2 + \beta_{36}O_2A_2 + \beta_{37}A_1A_2 + \beta_{38}O_3 \nonumber \\
&& + (\psi_0 + \psi_1 O_3 + \psi_2 A_2 + \psi_3 A_1A_2)A_3, \nonumber \\
Q_2 &=& \beta_{20} + \beta_{21}O_1 + \beta_{22}A_1 + \beta_{23}O_1A_1 + \beta_{24}O_2 + (\psi_0 + \psi_1 O_2 + \psi_2 A_1)A_2, \nonumber \\
Q_1 &=& \beta_{10} + \beta_{11}O_1 + (\psi_0 + \psi_1 O_1)A_1. \nonumber
\end{eqnarray}
Note that $\psi_0$ and $\psi_1$ are the shared parameters across all three stages, whereas $\psi_2$ is shared only between stages 2 and 3. To calculate the bias of different parameters, the relationship between $\gamma$s and $\beta$s are given in \cite{Chak_2016}. In Table \ref{Tab: allocation-match}, the extent of nonregularity (due to non-smooth maximization) is given by $p_3 = P(\psi^TH_{31} = 0)$ and $p_2 = P(\psi^TH_{21} = 0)$ corresponding to stages 3 and 2 which can cause bias in parameter estimation in stages 2 and 1, respectively.

We can observe from Table \ref{Tab: allocation-match} that the penalized Q-shared algorithm performs better in all the examples except in Ex. 4 and for the simple average (SA) version of the algorithms in Ex. 2 with respect to both metrics $M$ and $\Tilde{M}$. In some cases, the gain in allocation matching is high. For example, in Ex. 3, there is around a 3 to 4\% increase in allocation matching with respect to both metrics. Overall, we notice that the allocation matching does not depend on the choice of initial values for both the Q-shared and the penalized Q-shared algorithms. Interestingly, both algorithms perform well based on the initial values of all the parameters set to zero. Based on these observations, we can say that the developed penalized Q-shared algorithm is quite robust to the choice of initial values of the parameters. We have reported the bias of only $\psi_0$ as it represents the ``main effect" of treatment $A_j$ at the $j^{th}$ stage. 


\begin{tiny}
\begin{center}
	\begin{longtable}{lllcccccc}
		\caption{Comparison of Q-shared and penalized Q-shared algorithms with respect to the bias of $\psi_0$ and allocation matching with the oracle (\%) for sample size 300. The numbers for Q-shared have been recalculated and are the same as in \cite{Chak_2016}.} \label{Tab: allocation-match}\\
		\hline
		& & & \multicolumn{3}{c}{Q-Shared}  & \multicolumn{3}{c}{Penalized Q-Shared}\\ \cmidrule{4-6}  \cdashline{7-9}
		& & & & \multicolumn{2}{c}{\underline{Allocation Matching}}  & & \multicolumn{2}{c}{\underline{Allocation Matching}}\\ 
		Ex. & $(\psi_0,p_3,p_2)$ & Initial Values & Bias ($\psi_0$) & $\mathbf{M(\%)}$ &  $\mathbf{\Tilde{M}(\%)}$ & Bias ($\psi_0$) & $\mathbf{M(\%)}$ &  $\mathbf{\Tilde{M}(\%)}$\\ 
		\hline
		1 & (0.01,0,0) & SA & 0.0158 & 58.78 & 43.47 &   0.0269 & 62.73 & 49.00 \\ 
		&  & IVWA & 0.0154 & 59.08 & 43.74 &  0.0265 & 62.60 & 49.22  \\ 
		&  & MAX & 0.0156 & 59.23 & 43.90 &  0.0267 & 62.66 & 49.24  \\ 
		&  & MIN & 0.0154 & 59.14 & 43.80 &  0.0265 & 62.63 & 49.14  \\ 
		&  & ZERO & 0.0154 & 59.13 & 43.80 &  0.0266 & 62.71 & 49.28  \\
		&  &  &  &  &  &  &  &  \\ 
		2 & (-0.05,0.5,0.5) & SA & -0.0062 & 75.64 & 67.21 &  0.0016 & 75.22 & 66.76  \\ 
		&  & IVWA & -0.0063 & 75.57 & 67.11 &  0.0014 & 75.89 & 67.85   \\ 
		&  & MAX & -0.0061 & 75.52 & 67.15 &  0.0015 & 75.70 & 67.68  \\ 
		&  & MIN & -0.0063 & 75.64 & 67.20 &   0.0013 & 75.87 & 67.84  \\ 
		&  & ZERO & -0.0063 & 75.60 & 67.17 & 0.0014 & 75.73 & 67.67  \\
		&  &  &  &  & &  &  &   \\ 
		3 & (0.05,0.5,0) & SA & -0.0134 & 67.42 & 51.45  &  -0.0139 & 70.26 & 55.45 \\ 
		&  & IVWA & -0.0132 & 67.86 & 51.91  &  -0.0139 & 70.58 & 55.49  \\ 
		&  & MAX & -0.0129 & 67.96 & 52.07  &  -0.0138 & 70.65 & 55.52  \\ 
		&  & MIN & -0.0132 & 67.82 & 51.88  &  -0.0139 & 70.24 & 55.08  \\ 
		&  & ZERO & -0.0132 & 67.83 & 51.94  &  -0.0138 & 70.61 & 55.51 \\ 
		&  &  &  &  &  &  &  &   \\ 
		4 & (0.05,0,0) & SA & -0.0119 & 69.53 & 54.13  &  -0.0175 & 69.02 & 53.62  \\ 
		&  & IVWA & -0.0119 & 69.25 & 54.17  & -0.0172 & 68.37 & 52.89  \\ 
		&  & MAX & -0.0117 & 69.30 & 54.27  & -0.0171 & 68.36 & 52.85  \\ 
		&  & MIN & -0.0120 & 69.23 & 54.21  & -0.0173 & 68.34 & 52.86  \\ 
		&  & ZERO & -0.0119 & 69.17 & 54.21  & -0.0172 & 68.24 & 52.63 \\ 
		&  &  &  &  &   &  &  &  \\ 
		5 & (0.1,0.25,0) & SA & 0.0196 & 93.06 & 88.56  & 0.0069 & 94.67 & 91.01  \\ 
		&  & IVWA & 0.0199 & 93.29 & 88.96  & 0.0069 & 94.28 & 90.50  \\ 
		&  & MAX & 0.0201 & 93.38 & 89.08  & 0.0070 & 94.27 & 90.51  \\ 
		&  & MIN & 0.0199 & 93.27 & 88.97  & 0.0069 & 94.30 & 90.50  \\ 
		&  & ZERO & 0.0199 & 93.29 & 88.95  & 0.0069 & 94.28 & 90.48  \\ 
		&  &  &  &  &   &  &  &  \\ 
		6 & (0.1,0.25,0.25) & SA & 0.0046 & 87.12 & 80.52  & -0.0050 & 88.04 & 81.94 \\ 
		&  & IVWA & 0.0063 & 90.32 & 85.12  & -0.0037 & 91.27 & 86.44  \\ 
		&  & MAX & 0.0064 & 90.46 & 85.30  & -0.0036 & 91.29 & 86.45\\ 
		&  & MIN & 0.0063 & 90.43 & 85.25  & -0.0036 & 91.17 & 86.25  \\ 
		&  & ZERO & 0.0063 & 90.46 & 85.25  & -0.0036 & 91.32 & 86.47\\ 
		&  &  &  &  &   &  &  &  \\ 
		7 & (-0.1,0,0) & SA & -0.0055 & 89.26 & 79.65  & 0.0032 & 89.93 & 80.99  \\ 
		&  & IVWA & -0.0053 & 88.55 & 78.75  & 0.0029 & 90.07 & 81.23  \\ 
		&  & MAX & -0.0052 & 88.60 & 78.80  & 0.0030 & 90.04 & 81.20   \\ 
		&  & MIN & -0.0053 & 88.62 & 78.77  & 0.0029 & 90.06 & 81.23    \\ 
		&  & ZERO & -0.0052 & 88.65 & 78.81  & 0.0030 & 90.05 & 81.16  \\ 
		\hline
	\end{longtable}
\end{center}
\end{tiny}

\section{Application to Real data}
We have applied the penalized Q-shared algorithm to the three-stage STAR*D data to estimate the optimal shared-parameter DTR in the context of a major depressive disorder. A detailed description of the STAR*D trial that consists of 4041 patients is available in Rush et al. \cite{rush04}. For simplicity, here we consider binary treatment ($A_j$) at each stage ($j=1,2,3$) as mono therapy (coded as 1) and combination therapy (coded as -1), and other setup is also the same as Chakraborty et al. \cite{Chak_2016}. The primary outcome ($Y_{primary}$) is the same as defined in Section \ref{Non-covg_Q-Pena} as a function of stage-specific outcomes $Y_1, Y_2$ and $Y_3$ and response indicators ($R_1, R_2$). Here, $Y_j, j=1,2,3$ denotes $-$QIDS (rated by a clinician). The negative sign before QIDS makes the outcome such that the higher the better since a lower QIDS score indicates an improvement in a patient outcome. Therefore, $Y_{primary}$ denotes an average $-$QIDS score of an individual during their stay in the trial. We consider four tailoring variables (covariates), initial QIDS score at the start of a stage ($start.QIDS_j$), observed slope of the QIDS score from the previous stage ($slope.QIDS_j$), presence or absence of side effect from the earlier stage ($side.effect_j$) and earlier stage binary treatment ($A_{j-1} \in \{-1, 1\}$, j=2,3). We consider the following Q-functions \citep{Chak_2016}: 
\begin{small}
\begin{eqnarray}
Q_3(H_3, A_3) &=& \beta_{03} + \beta_{13}\times start.QIDS_3 + \beta_{23}\times slope.QIDS_3 + \beta_{33}\times side.effect_3 + \beta_{43}\times A_2 \nonumber \\ 
&& + (\psi_0 + \psi_1\times start.QIDS_3 + \psi_2\times slope.QIDS_3 + \psi_3\times side.effect_3 + \psi_4\times A_2) \times A_3 \nonumber \\ 
Q_2(H_2, A_2) &=& \beta_{02} + \beta_{12}\times start.QIDS_2 + \beta_{22}\times slope.QIDS_2 + \beta_{32}\times side.effect_2 + \beta_{42}\times A_1 \nonumber \\ 
&& + (\psi_0 + \psi_1\times start.QIDS_2 + \psi_2\times slope.QIDS_2 + \psi_3\times side.effect_2 + \psi_4\times A_1) \times A_2 \nonumber \\ 
Q_2(H_1, A_1) &=& \beta_{01} + \beta_{11}\times start.QIDS_1 + \beta_{21}\times slope.QIDS_1 \nonumber \\ 
&& + (\psi_0 + \psi_1\times start.QIDS_1 + \psi_2\times slope.QIDS_1 ) \times A_1 \nonumber
\end{eqnarray}
\end{small}
Note that the side effects from the previous stage for stage 1 are not considered, as everyone obtained the same treatment (citalopram) before entering stage 1. Table \ref{Tab_SATRD:Qshared} shows the estimated shared parameters using penalized Q-shared algorithm along with $m$-out-of-$n$ bootstrap variances and 95\% confidence intervals. Observe that the only significant shared parameter is $\psi_2$. This finding from STAR*D data is consistent with that using the Q-shared algorithm mentioned in \cite{Chak_2016}. Using the estimated values of the shared parameters, the stage-specific optimal decision rules are
\begin{small}
\begin{eqnarray}
d_3(H_3) &=& sign(-0.0298 + 0.0052\times start.QIDS_3 -0.0700\times slope.QIDS_3 \nonumber \\
&& \hspace{4cm} + 0.0086\times side.effect_3 + 0.0060\times A_2), \nonumber \\
d_2(H_2) &=& sign(-0.0298 + 0.0052\times start.QIDS_2 -0.0700\times slope.QIDS_2 \nonumber \\
&& \hspace{4cm} + 0.0086\times side.effect_2 + 0.0060\times A_1), \nonumber \\
d_1(H_1) &=& sign(-0.0298 + 0.0052\times start.QIDS_1 -0.0700\times slope.QIDS_1 ). \nonumber 
\end{eqnarray}
\end{small}
In practice, we will know the stage-specific values of $start.QIDS$, $slope.QIDS$, $side.effect$, and the previous stage treatment for a patient. Therefore, using the above DTRs, one can decide whether to give mono therapy (1) or combination therapy (-1) to that patient in a particular stage.

\begin{table}[ht]
	\centering
	\caption{Estimates of the shared parameters along with their $m$-out-of-$n$ bootstrap variances and confidence intervals generated by penalized Q-shared algorithm using the STAR*D data. Here the binary treatment is coded as \{-1, 1\}.}
	\label{Tab_SATRD:Qshared}
	\begin{tabular}{crrcc}
		\hline
		& & \multicolumn{2}{c}{ Using $m$-out-of-$n$ bootstrap} \\ \cmidrule{3-4}
		Shared Parameter & Estimate & Variance & CI \\ 
		\hline
		$\psi_0$ & -0.0298  & 0.0010 &	(-0.0901, 0.0290) \\
		$\psi_1$ &  0.0052  & 8.6309 &	(-0.0141, 0.0220) \\
		$\psi_2$ & -0.0700 & 0.0014 &	(-0.1378, -0.0001) \\
		$\psi_3$ & 0.0086   & 0.0007 &	(-0.0436, 0.0584) \\
		$\psi_4$ & 0.0060   & 0.0017 &	(-0.0786, 0.0876) \\
		\hline
	\end{tabular}
\end{table}


\section{Discussion}

In some clinical practices, consistent decision rules across different disease stages are desired to provide optimal treatment. The Q-shared algorithm is a Q-learning procedure with shared parameters among linear regression models across stages, enabling such shared decision-making. The Q-shared algorithm estimates fewer parameters than a Q-learning algorithm to develop DTRs based on the same data for shared decision rules. However, as shown in Section~\ref{Non-covg_Q-Pena}, the original Q-shared algorithm may fail to converge in certain settings. To address this issue, we have proposed the penalized Q-shared algorithm in the current article. 
Our simulation study demonstrates that the penalized Q-shared algorithm outperforms the original Q-shared algorithm in terms of correctly allocating patients to the true optimal decision rules, even in settings without convergence issues. Therefore, researchers may consider using the penalized Q-shared algorithm over the original Q-shared algorithm when developing shared decision rules is the primary research objective.

This paper empirically demonstrates the improvements achieved by the penalized Q-shared algorithm over the original Q-shared algorithm using comprehensive simulation studies, and illustrates the use of the penalized Q-shared approach in a real application. Establishing theoretical guarantees of superiority using penalized Q-shared remains for future work. Additionally, this paper only focuses on binary treatment options at each of the three stages. However, it is straightforward to extend the penalized Q-shared algorithm to accommodate settings with multiple treatment options and additional stages. Furthermore, tree-based or list-based DTRs provide interpretable and clinically meaningful decision rules \citep{zhang2018interpretable, zhou2023estimating}. Future work may focus on developing shared decision rules for these types of DTRs.

\bmhead{Acknowledgements}
Trikay Nalamada and Shruti Agarwal acknowledge support from Samsung Fellowship. Bibhas Chakraborty acknowledges support from the start-up grant from the Duke-NUS Medical School, Singapore. Palash Ghosh acknowledges support from the Start-up Grant (PG001), Indian Institute of Technology Guwahati. 

\section*{Declarations}

\textbf{Conflict of interest}: The authors declare that they have no conflict of interest.

\textbf{Data Availability}: The data are not publicly available due to privacy and ethical restrictions.






\bibliography{refs}


\begin{thebibliography}{32}
\ifx \bisbn   \undefined \def \bisbn  #1{ISBN #1}\fi
\ifx \binits  \undefined \def \binits#1{#1}\fi
\ifx \bauthor  \undefined \def \bauthor#1{#1}\fi
\ifx \batitle  \undefined \def \batitle#1{#1}\fi
\ifx \bjtitle  \undefined \def \bjtitle#1{#1}\fi
\ifx \bvolume  \undefined \def \bvolume#1{\textbf{#1}}\fi
\ifx \byear  \undefined \def \byear#1{#1}\fi
\ifx \bissue  \undefined \def \bissue#1{#1}\fi
\ifx \bfpage  \undefined \def \bfpage#1{#1}\fi
\ifx \blpage  \undefined \def \blpage #1{#1}\fi
\ifx \burl  \undefined \def \burl#1{\textsf{#1}}\fi
\ifx \doiurl  \undefined \def \doiurl#1{\url{https://doi.org/#1}}\fi
\ifx \betal  \undefined \def \betal{\textit{et al.}}\fi
\ifx \binstitute  \undefined \def \binstitute#1{#1}\fi
\ifx \binstitutionaled  \undefined \def \binstitutionaled#1{#1}\fi
\ifx \bctitle  \undefined \def \bctitle#1{#1}\fi
\ifx \beditor  \undefined \def \beditor#1{#1}\fi
\ifx \bpublisher  \undefined \def \bpublisher#1{#1}\fi
\ifx \bbtitle  \undefined \def \bbtitle#1{#1}\fi
\ifx \bedition  \undefined \def \bedition#1{#1}\fi
\ifx \bseriesno  \undefined \def \bseriesno#1{#1}\fi
\ifx \blocation  \undefined \def \blocation#1{#1}\fi
\ifx \bsertitle  \undefined \def \bsertitle#1{#1}\fi
\ifx \bsnm \undefined \def \bsnm#1{#1}\fi
\ifx \bsuffix \undefined \def \bsuffix#1{#1}\fi
\ifx \bparticle \undefined \def \bparticle#1{#1}\fi
\ifx \barticle \undefined \def \barticle#1{#1}\fi
\bibcommenthead
\ifx \bconfdate \undefined \def \bconfdate #1{#1}\fi
\ifx \botherref \undefined \def \botherref #1{#1}\fi
\ifx \url \undefined \def \url#1{\textsf{#1}}\fi
\ifx \bchapter \undefined \def \bchapter#1{#1}\fi
\ifx \bbook \undefined \def \bbook#1{#1}\fi
\ifx \bcomment \undefined \def \bcomment#1{#1}\fi
\ifx \oauthor \undefined \def \oauthor#1{#1}\fi
\ifx \citeauthoryear \undefined \def \citeauthoryear#1{#1}\fi
\ifx \endbibitem  \undefined \def \endbibitem {}\fi
\ifx \bconflocation  \undefined \def \bconflocation#1{#1}\fi
\ifx \arxivurl  \undefined \def \arxivurl#1{\textsf{#1}}\fi
\csname PreBibitemsHook\endcsname

\bibitem[\protect\citeauthoryear{Murphy}{2003}]{murphy03}
\begin{barticle}
\bauthor{\bsnm{Murphy}, \binits{S.A.}}:
\batitle{Optimal dynamic treatment regimes (with discussions)}.
\bjtitle{Journal of the Royal Statistical Society, Series B}
\bvolume{65},
\bfpage{331}--\blpage{366}
(\byear{2003})
\end{barticle}
\endbibitem

\bibitem[\protect\citeauthoryear{Robins}{2004}]{robins04}
\begin{bchapter}
\bauthor{\bsnm{Robins}, \binits{J.M.}}:
\bctitle{Optimal structural nested models for optimal sequential decisions}.
In: \beditor{\bsnm{Lin}, \binits{D.Y.}},
\beditor{\bsnm{Heagerty}, \binits{P.}} (eds.)
\bbtitle{Proceedings of the Second Seattle Symposium on Biostatistics},
pp. \bfpage{189}--\blpage{326}.
\bpublisher{Springer},
\blocation{New York}
(\byear{2004})
\end{bchapter}
\endbibitem

\bibitem[\protect\citeauthoryear{Chakraborty and Moodie}{2013}]{Chak_book}
\begin{bbook}
\bauthor{\bsnm{Chakraborty}, \binits{B.}},
\bauthor{\bsnm{Moodie}, \binits{E.E.M.}}:
\bbtitle{Statistical Methods for Dynamic Treatment Regimes: Reinforcement Learning, Causal Inference, and Personalized Medicine}.
\bpublisher{Springer},
\blocation{New York}
(\byear{2013})
\end{bbook}
\endbibitem

\bibitem[\protect\citeauthoryear{Ghosh et~al.}{2020}]{Ghosh2020noninfi}
\begin{barticle}
\bauthor{\bsnm{Ghosh}, \binits{P.}},
\bauthor{\bsnm{Nahum-Shani}, \binits{I.}},
\bauthor{\bsnm{Spring}, \binits{B.}},
\bauthor{\bsnm{Chakraborty}, \binits{B.}}:
\batitle{Noninferiority and equivalence tests in sequential, multiple assignment, randomized trials ({SMARTs}).}
\bjtitle{Psychological Methods}
\bvolume{25}(\bissue{2}),
\bfpage{182}
(\byear{2020})
\end{barticle}
\endbibitem

\bibitem[\protect\citeauthoryear{Nahum-Shani et~al.}{2015}]{nahum2015Health_psycho}
\begin{barticle}
\bauthor{\bsnm{Nahum-Shani}, \binits{I.}},
\bauthor{\bsnm{Hekler}, \binits{E.B.}},
\bauthor{\bsnm{Spruijt-Metz}, \binits{D.}}:
\batitle{Building health behavior models to guide the development of just-in-time adaptive interventions: A pragmatic framework.}
\bjtitle{Health Psychology}
\bvolume{34}(\bissue{S}),
\bfpage{1209}
(\byear{2015})
\end{barticle}
\endbibitem

\bibitem[\protect\citeauthoryear{Eden}{2017}]{eden2017field}
\begin{barticle}
\bauthor{\bsnm{Eden}, \binits{D.}}:
\batitle{Field experiments in organizations}.
\bjtitle{Annual Review of Organizational Psychology and Organizational Behavior}
\bvolume{4},
\bfpage{91}--\blpage{122}
(\byear{2017})
\end{barticle}
\endbibitem

\bibitem[\protect\citeauthoryear{Song et~al.}{2015}]{song2015sparse}
\begin{barticle}
\bauthor{\bsnm{Song}, \binits{R.}},
\bauthor{\bsnm{Kosorok}, \binits{M.}},
\bauthor{\bsnm{Zeng}, \binits{D.}},
\bauthor{\bsnm{Zhao}, \binits{Y.}},
\bauthor{\bsnm{Laber}, \binits{E.}},
\bauthor{\bsnm{Yuan}, \binits{M.}}:
\batitle{On sparse representation for optimal individualized treatment selection with penalized outcome weighted learning}.
\bjtitle{STAT}
\bvolume{4}(\bissue{1}),
\bfpage{59}--\blpage{68}
(\byear{2015})
\end{barticle}
\endbibitem

\bibitem[\protect\citeauthoryear{Zhang et~al.}{2012}]{Zhang12a}
\begin{barticle}
\bauthor{\bsnm{Zhang}, \binits{B.}},
\bauthor{\bsnm{Tsiatis}, \binits{A.A.}},
\bauthor{\bsnm{Davidian}, \binits{M.}},
\bauthor{\bsnm{Zhang}, \binits{M.}},
\bauthor{\bsnm{Laber}, \binits{E.B.}}:
\batitle{Estimating optimal treatment regimes from a classification perspective}.
\bjtitle{STAT}
\bvolume{1},
\bfpage{103}--\blpage{114}
(\byear{2012})
\end{barticle}
\endbibitem

\bibitem[\protect\citeauthoryear{Kidwell}{2014}]{kidwell2014smart}
\begin{barticle}
\bauthor{\bsnm{Kidwell}, \binits{K.M.}}:
\batitle{{SMART} designs in cancer research: past, present, and future}.
\bjtitle{Clinical Trials}
\bvolume{11}(\bissue{4}),
\bfpage{445}--\blpage{456}
(\byear{2014})
\end{barticle}
\endbibitem

\bibitem[\protect\citeauthoryear{Almirall et~al.}{2016}]{almirall2016Autism}
\begin{barticle}
\bauthor{\bsnm{Almirall}, \binits{D.}},
\bauthor{\bsnm{DiStefano}, \binits{C.}},
\bauthor{\bsnm{Chang}, \binits{Y.-C.}},
\bauthor{\bsnm{Shire}, \binits{S.}},
\bauthor{\bsnm{Kaiser}, \binits{A.}},
\bauthor{\bsnm{Lu}, \binits{X.}},
\bauthor{\bsnm{Nahum-Shani}, \binits{I.}},
\bauthor{\bsnm{Landa}, \binits{R.}},
\bauthor{\bsnm{Mathy}, \binits{P.}},
\bauthor{\bsnm{Kasari}, \binits{C.}}:
\batitle{Longitudinal effects of adaptive interventions with a speech-generating device in minimally verbal children with asd}.
\bjtitle{Journal of Clinical Child \& Adolescent Psychology}
\bvolume{45}(\bissue{4}),
\bfpage{442}--\blpage{456}
(\byear{2016})
\end{barticle}
\endbibitem

\bibitem[\protect\citeauthoryear{Almirall et~al.}{2014}]{Almirall2014}
\begin{barticle}
\bauthor{\bsnm{Almirall}, \binits{D.}},
\bauthor{\bsnm{Nahum-Shani}, \binits{I.}},
\bauthor{\bsnm{Sherwood}, \binits{N.E.}},
\bauthor{\bsnm{Murphy}, \binits{S.A.}}:
\batitle{Introduction to {SMART} designs for the development of adaptive interventions: with application to weight loss research}.
\bjtitle{Translational Behavioral Medicine}
\bvolume{4}(\bissue{3}),
\bfpage{260}--\blpage{274}
(\byear{2014})
\doiurl{10.1007/s13142-014-0265-0}
\end{barticle}
\endbibitem

\bibitem[\protect\citeauthoryear{Walkington}{2013}]{Walkington2013Education}
\begin{barticle}
\bauthor{\bsnm{Walkington}, \binits{C.A.}}:
\batitle{Using adaptive learning technologies to personalize instruction to student interests: The impact of relevant contexts on performance and learning outcomes.}
\bjtitle{Journal of Educational Psychology}
\bvolume{105}(\bissue{4}),
\bfpage{932}
(\byear{2013})
\end{barticle}
\endbibitem

\bibitem[\protect\citeauthoryear{Rush et~al.}{2003}]{rush2003QUID}
\begin{barticle}
\bauthor{\bsnm{Rush}, \binits{A.J.}},
\bauthor{\bsnm{Trivedi}, \binits{M.H.}},
\bauthor{\bsnm{Ibrahim}, \binits{H.M.}},
\bauthor{\bsnm{Carmody}, \binits{T.J.}},
\bauthor{\bsnm{Arnow}, \binits{B.}},
\bauthor{\bsnm{Klein}, \binits{D.N.}},
\bauthor{\bsnm{Markowitz}, \binits{J.C.}},
\bauthor{\bsnm{Ninan}, \binits{P.T.}},
\bauthor{\bsnm{Kornstein}, \binits{S.}},
\bauthor{\bsnm{Manber}, \binits{R.}}, \betal:
\batitle{The 16-item quick inventory of depressive symptomatology (qids), clinician rating (qids-c), and self-report (qids-sr): a psychometric evaluation in patients with chronic major depression}.
\bjtitle{Biological Psychiatry}
\bvolume{54}(\bissue{5}),
\bfpage{573}--\blpage{583}
(\byear{2003})
\end{barticle}
\endbibitem

\bibitem[\protect\citeauthoryear{Villain et~al.}{2019}]{villain2019adaptive}
\begin{barticle}
\bauthor{\bsnm{Villain}, \binits{L.}},
\bauthor{\bsnm{Commenges}, \binits{D.}},
\bauthor{\bsnm{Pasin}, \binits{C.}},
\bauthor{\bsnm{Prague}, \binits{M.}},
\bauthor{\bsnm{Thi{\'e}baut}, \binits{R.}}:
\batitle{Adaptive protocols based on predictions from a mechanistic model of the effect of {IL7} on {CD4} counts}.
\bjtitle{Statistics in Medicine}
\bvolume{38}(\bissue{2}),
\bfpage{221}--\blpage{235}
(\byear{2019})
\end{barticle}
\endbibitem

\bibitem[\protect\citeauthoryear{{TIE Committee}}{2009}]{international2009international}
\begin{barticle}
\bauthor{\bsnm{{TIE Committee}}}:
\batitle{International expert committee report on the role of the {A1c} assay in the diagnosis of diabetes}.
\bjtitle{Diabetes Care}
\bvolume{32}(\bissue{7}),
\bfpage{1327}--\blpage{1334}
(\byear{2009})
\end{barticle}
\endbibitem

\bibitem[\protect\citeauthoryear{Zhao et~al.}{2020}]{zhao2020constructing}
\begin{barticle}
\bauthor{\bsnm{Zhao}, \binits{Y.-Q.}},
\bauthor{\bsnm{Zhu}, \binits{R.}},
\bauthor{\bsnm{Chen}, \binits{G.}},
\bauthor{\bsnm{Zheng}, \binits{Y.}}:
\batitle{Constructing dynamic treatment regimes with shared parameters for censored data}.
\bjtitle{Statistics in Medicine}
\bvolume{39}(\bissue{9}),
\bfpage{1250}--\blpage{1263}
(\byear{2020})
\end{barticle}
\endbibitem

\bibitem[\protect\citeauthoryear{Chakraborty et~al.}{2016}]{Chak_2016}
\begin{barticle}
\bauthor{\bsnm{Chakraborty}, \binits{B.}},
\bauthor{\bsnm{Ghosh}, \binits{P.}},
\bauthor{\bsnm{Moodie}, \binits{E.E.M.}},
\bauthor{\bsnm{Rush}, \binits{A.J.}}:
\batitle{Estimating optimal shared-parameter dynamic regimens with application to a multistage depression clinical trial}.
\bjtitle{Biometrics}
\bvolume{72}(\bissue{3}),
\bfpage{865}--\blpage{876}
(\byear{2016})
\doiurl{10.1111/biom.12493}
\end{barticle}
\endbibitem

\bibitem[\protect\citeauthoryear{Song et~al.}{2015}]{RuiSong2015}
\begin{barticle}
\bauthor{\bsnm{Song}, \binits{R.}},
\bauthor{\bsnm{Wang}, \binits{W.}},
\bauthor{\bsnm{Zeng}, \binits{D.}},
\bauthor{\bsnm{Kosorok}, \binits{M.R.}}:
\batitle{Penalized {Q}-learning for dynamic treatment regimens}.
\bjtitle{Statistica Sinica}
\bvolume{25}(\bissue{3}),
\bfpage{901}--\blpage{920}
(\byear{2015})
\end{barticle}
\endbibitem

\bibitem[\protect\citeauthoryear{Zhao et~al.}{2020}]{zhao2020Shared_Censored}
\begin{barticle}
\bauthor{\bsnm{Zhao}, \binits{Y.-Q.}},
\bauthor{\bsnm{Zhu}, \binits{R.}},
\bauthor{\bsnm{Chen}, \binits{G.}},
\bauthor{\bsnm{Zheng}, \binits{Y.}}:
\batitle{Constructing dynamic treatment regimes with shared parameters for censored data}.
\bjtitle{Statistics in Medicine}
\bvolume{39}(\bissue{9}),
\bfpage{1250}--\blpage{1263}
(\byear{2020})
\end{barticle}
\endbibitem

\bibitem[\protect\citeauthoryear{Schulte et~al.}{2014}]{schulte2014q}
\begin{barticle}
\bauthor{\bsnm{Schulte}, \binits{P.J.}},
\bauthor{\bsnm{Tsiatis}, \binits{A.A.}},
\bauthor{\bsnm{Laber}, \binits{E.B.}},
\bauthor{\bsnm{Davidian}, \binits{M.}}:
\batitle{Q-and {A}-learning methods for estimating optimal dynamic treatment regimes}.
\bjtitle{Statistical Science}
\bvolume{29}(\bissue{4}),
\bfpage{640}
(\byear{2014})
\end{barticle}
\endbibitem

\bibitem[\protect\citeauthoryear{{van der Laan} and Rubin}{2006}]{vanderLaanRubin2006}
\begin{botherref}
\oauthor{\bsnm{{van der Laan}}, \binits{M.J.}},
\oauthor{\bsnm{Rubin}, \binits{D.}}:
Targeted maximum likelihood learning.
The International Journal of Biostatistics
\textbf{2}(1)
(2006)
\doiurl{10.2202/1557-4679.1043}
\end{botherref}
\endbibitem

\bibitem[\protect\citeauthoryear{Ertefaie et~al.}{2021}]{ertefaie2021robust}
\begin{barticle}
\bauthor{\bsnm{Ertefaie}, \binits{A.}},
\bauthor{\bsnm{McKay}, \binits{J.R.}},
\bauthor{\bsnm{Oslin}, \binits{D.}},
\bauthor{\bsnm{Strawderman}, \binits{R.L.}}:
\batitle{Robust {Q}-learning}.
\bjtitle{Journal of the American Statistical Association}
\bvolume{116}(\bissue{533}),
\bfpage{368}--\blpage{381}
(\byear{2021})
\end{barticle}
\endbibitem

\bibitem[\protect\citeauthoryear{Tsiatis et~al.}{2019}]{tsiatis2019dynamic}
\begin{bbook}
\bauthor{\bsnm{Tsiatis}, \binits{A.A.}},
\bauthor{\bsnm{Davidian}, \binits{M.}},
\bauthor{\bsnm{Holloway}, \binits{S.T.}},
\bauthor{\bsnm{Laber}, \binits{E.B.}}:
\bbtitle{Dynamic Treatment Regimes: Statistical Methods for Precision Medicine}.
\bpublisher{Chapman and Hall/CRC},
\blocation{1st edition}
(\byear{2019})
\end{bbook}
\endbibitem

\bibitem[\protect\citeauthoryear{Zhang et~al.}{2018}]{zhang2018interpretable}
\begin{barticle}
\bauthor{\bsnm{Zhang}, \binits{Y.}},
\bauthor{\bsnm{Laber}, \binits{E.B.}},
\bauthor{\bsnm{Davidian}, \binits{M.}},
\bauthor{\bsnm{Tsiatis}, \binits{A.A.}}:
\batitle{Interpretable dynamic treatment regimes}.
\bjtitle{Journal of the American Statistical Association}
\bvolume{113}(\bissue{524}),
\bfpage{1541}--\blpage{1549}
(\byear{2018})
\end{barticle}
\endbibitem

\bibitem[\protect\citeauthoryear{Zhou et~al.}{2023}]{zhou2023estimating}
\begin{barticle}
\bauthor{\bsnm{Zhou}, \binits{N.}},
\bauthor{\bsnm{Wang}, \binits{L.}},
\bauthor{\bsnm{Almirall}, \binits{D.}}:
\batitle{Estimating tree-based dynamic treatment regimes using observational data with restricted treatment sequences}.
\bjtitle{Biometrics}
\bvolume{79}(\bissue{3}),
\bfpage{2260}--\blpage{2271}
(\byear{2023})
\end{barticle}
\endbibitem

\bibitem[\protect\citeauthoryear{Murphy}{2005}]{murphy05b}
\begin{barticle}
\bauthor{\bsnm{Murphy}, \binits{S.A.}}:
\batitle{A generalization error for {Q}-learning}.
\bjtitle{Journal of Machine Learning Research}
\bvolume{6},
\bfpage{1073}--\blpage{1097}
(\byear{2005})
\end{barticle}
\endbibitem

\bibitem[\protect\citeauthoryear{Lizotte}{2011}]{lizotte2011convergent}
\begin{barticle}
\bauthor{\bsnm{Lizotte}, \binits{D.}}:
\batitle{Convergent fitted value iteration with linear function approximation}.
\bjtitle{Advances in Neural Information Processing Systems}
\bvolume{24},
\bfpage{2537}--\blpage{2545}
(\byear{2011})
\end{barticle}
\endbibitem

\bibitem[\protect\citeauthoryear{Gordon}{1999}]{gordon1999approximate}
\begin{bbook}
\bauthor{\bsnm{Gordon}, \binits{G.J.}}:
\bbtitle{Approximate Solutions to Markov Decision Processes}.
\bpublisher{Carnegie Mellon University},
\blocation{PhD Thesis}
(\byear{1999})
\end{bbook}
\endbibitem

\bibitem[\protect\citeauthoryear{Chakraborty et~al.}{2013}]{chak_laber12}
\begin{barticle}
\bauthor{\bsnm{Chakraborty}, \binits{B.}},
\bauthor{\bsnm{Laber}, \binits{E.B.}},
\bauthor{\bsnm{Zhao}, \binits{Y.}}:
\batitle{Inference for optimal dynamic treatment regimes using an adaptive $m$-out-of-$n$ bootstrap scheme}.
\bjtitle{Biometrics}
\bvolume{69},
\bfpage{714}--\blpage{723}
(\byear{2013})
\end{barticle}
\endbibitem

\bibitem[\protect\citeauthoryear{Hastie et~al.}{2017}]{hastie2009elements}
\begin{bbook}
\bauthor{\bsnm{Hastie}, \binits{T.}},
\bauthor{\bsnm{Tibshirani}, \binits{R.}},
\bauthor{\bsnm{Friedman}, \binits{J.}}:
\bbtitle{The Elements of Statistical Learning: Data Mining, Inference, and Prediction}.
\bpublisher{Springer},
\blocation{2nd edition}
(\byear{2017})
\end{bbook}
\endbibitem

\bibitem[\protect\citeauthoryear{Friedman et~al.}{2010}]{friedman2010regularization}
\begin{barticle}
\bauthor{\bsnm{Friedman}, \binits{J.}},
\bauthor{\bsnm{Hastie}, \binits{T.}},
\bauthor{\bsnm{Tibshirani}, \binits{R.}}:
\batitle{Regularization paths for generalized linear models via coordinate descent}.
\bjtitle{Journal of Statistical Software}
\bvolume{33}(\bissue{1}),
\bfpage{1}
(\byear{2010})
\end{barticle}
\endbibitem

\bibitem[\protect\citeauthoryear{Rush et~al.}{2004}]{rush04}
\begin{barticle}
\bauthor{\bsnm{Rush}, \binits{A.J.}},
\bauthor{\bsnm{Fava}, \binits{M.}},
\bauthor{\bsnm{Wisniewski}, \binits{S.R.}},
\bauthor{\bsnm{Lavori}, \binits{P.W.}},
\bauthor{\bsnm{Trivedi}, \binits{M.H.}},
\bauthor{\bsnm{Sackeim}, \binits{H.A.}},
\bauthor{\bsnm{Thase}, \binits{M.E.}},
\bauthor{\bsnm{Nierenberg}, \binits{A.A.}},
\bauthor{\bsnm{Quitkin}, \binits{F.M.}},
\bauthor{\bsnm{Kashner}, \binits{T.M.}},
\bauthor{\bsnm{Kupfer}, \binits{D.J.}},
\bauthor{\bsnm{Rosenbaum}, \binits{J.F.}},
\bauthor{\bsnm{Alpert}, \binits{J.}},
\bauthor{\bsnm{Stewart}, \binits{J.W.}},
\bauthor{\bsnm{McGrath}, \binits{P.J.}},
\bauthor{\bsnm{Biggs}, \binits{M.M.}},
\bauthor{\bsnm{Shores-Wilson}, \binits{K.}},
\bauthor{\bsnm{Lebowitz}, \binits{B.D.}},
\bauthor{\bsnm{Ritz}, \binits{L.}},
\bauthor{\bsnm{Niederehe}, \binits{G.}}:
\batitle{Sequenced treatment alternatives to relieve depression ({STAR*D}): {R}ationale and design}.
\bjtitle{Controlled Clinical Trials}
\bvolume{25},
\bfpage{119}--\blpage{142}
(\byear{2004})
\end{barticle}
\endbibitem

\end{thebibliography}

\backmatter

\end{document}